\pdfoutput=1

\documentclass[11pt]{article}

\usepackage{graphicx}
\usepackage{booktabs}

\usepackage{EMNLP2023}

\usepackage{times}
\usepackage{latexsym}

\usepackage[T1]{fontenc}

\usepackage[utf8]{inputenc}

\usepackage{microtype}

\usepackage{inconsolata}

\title{Flexible Model Interpretability through Natural Language Model Editing}

\author{Karel D'Oosterlinck$^1$, Thomas Demeester$^1$, Chris Develder$^1$, Christopher Potts$^2$ \\ $^1$Ghent University -- imec\qquad $^2$Stanford University \\
\texttt{karel.doosterlinck@ugent.be}}

\begin{document}
\maketitle

\section{Introduction}

Model interpretability and model editing are crucial goals in the age of large language models. 
Interestingly, there exists a link between these two goals: if a method is able to systematically edit model behavior with regard to a human concept of interest, this editor method can help make internal representations more interpretable by pointing towards relevant representations and systematically manipulating them.

This insight can be used to alleviate a limitation of existing explainability methods: learning how to faithfully understand and manipulate hidden representations with regard to a human-interpretable concept requires task-specific data and experiments \citep{vig2020causalmediation, geiger2021causal, pmlr-v162-geiger22a, meng2022locating, de2022sparse, olsson2022context}. Thus, these methods often struggle at the scale of our most widely-used models (\citealt{superalignment}; but see \citealt{Wu:Geiger:2023:BDAS}).

We propose to learn how to edit a model based on a natural language description of the edit, using generic instruction-tuning data. Crucially, we regularize these edits (e.g.\ restrict them to sparse interventions, to specific layers or to low-rank weight updates) such that they lead to some level of model understanding. The editing performance of different regularization approaches will highlight how faithful these assumptions are with regard to the model internals, across a broad range of concepts.

Other model editing work represents edits as input--output pairs \citep{mitchell2021fast, mitchell2022memory} and thus requires task-specific data to perform inference-time edits. If our proposed natural language editing generalizes to unseen instructions, it will provide significantly more flexibility at inference-time to perform task-specific edits and pursue new interpretability goals.

\begin{figure}
    \centering
    \includegraphics[width=\linewidth]{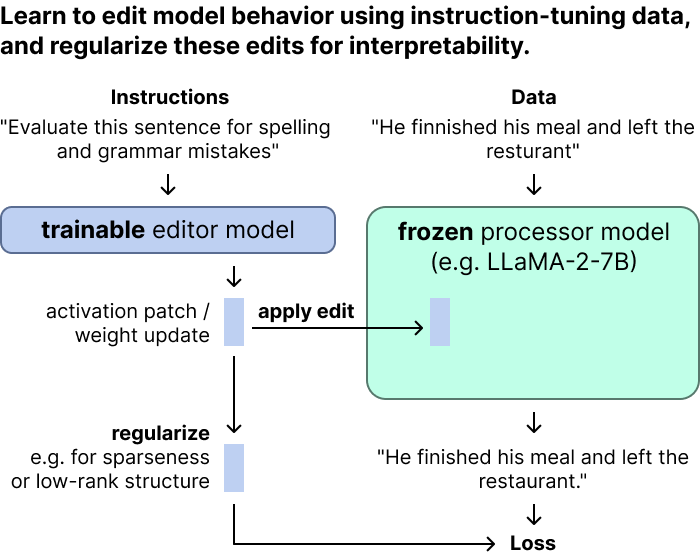}
    \caption{We train an editor model to perform regularized edits (e.g.\ sparse interventions) on a frozen processor model, given an instruction. Such an editor can be a flexible resource for downstream interpretability work. For example, if an editor learns to sparsely manipulate a frozen model, these sparse edits can teach us where information was localized in the original model.}
    \label{fig:editor-schematic}
    \vspace{-.5em}
\end{figure}

In this extended abstract, we report proof-of-concept results on learning to edit a model based on natural language instructions, without any edit regularization. We train a \texttt{GPT-2} \citep{radford2019language} \emph{editor} model to first process a generic natural language instruction (which typically conveys a human interpretable concept such as ``evaluate the following paragraph for spelling mistakes'') and then edit the forward pass of a frozen, 7 billion parameter \texttt{LLaMA-2} \citep{touvron2023llama} \emph{processor} model to manipulate its behavior on an input according to these instruction, as illustrated in Figure \ref{fig:editor-schematic}. These edits can be parameterized in many ways. For now we let \texttt{GPT-2} generate a new representation that we sum with a hidden representation of the \emph{processor} at a specific location. Intuitively, the \emph{editor} learns to efficiently inject the information represented in the instruction to systematically manipulate the \emph{processor}'s behavior. We set an empirical lower and upper bound for the editor's performance and achieve meaningful traction on this task, although a lot of headroom remains.

Editing a model based on natural language descriptions is valuable in and off itself, but does not yet convey any interpretability benefits. As a next step in our ongoing work, we plan to study the structure of model internals by parameterizing the editing procedure using different inductive biases. For example, one research area of explainable AI investigates whether human-interpretable concepts are localized in specific neurons or latent space directions. To study this, edits can be regularized to favor these localized manipulations. The resulting editor performance is indicative of whether the latent space actually had such a structure and the editor was able to systematically learn to manipulate this. If the editor performs well, its predicted edits could be used for downstream model editing or model explainability research.

\section{Methods and Results}
Consider a dataset $\mathcal{D}$ whose instances consist of an instruction $x_{i}$, a data input $x_{d}$, and a target $y$ (all natural language), an editor model $\mathcal{E}$, and a processor model $\mathcal{P}$. We denote a hidden representation of interest, formed during the forward pass of the processor as $h$. Given the instruction input and the hidden representation, the editor forms a new representation $\mathcal{E}\left(x_{i},h\right)$. This new representation replaces the original hidden representation $h$ during the forward pass of the processor on the data input $\mathcal{P}_{h \leftarrow \mathcal{E}\left(x_{i}, h\right)}\left(x_{d}\right)$. 

We define a loss $\mathcal{L}$ between the output of this manipulated forward pass and the target, and use its gradients $\nabla_\mathcal{E}\left( \mathcal{L}\left(\mathcal{P}_{h \leftarrow \mathcal{E}\left(x_{i}, h\right)}\left(x_{d}\right), y\right)\right)$ to optimize only the editor, keeping the processor $\mathcal{P}$ frozen. Intuitively, the editor is trained to inject information about the instruction $x_{i}$ into the forward pass of the processor in a manner that systematically changes the processor's behavior. 

In our experiments, a frozen \texttt{LLaMA-2-7B} acts as processor. The editor first maps the instruction $x_{i}$ to a latent vector using a trainable \texttt{GPT-2} model. Then, this vector is simply summed with the latent representation of the 1st token at layer \emph{l} of the processor's forward pass.

We consider a conceptual best and worst case bound to situate the editor performance. The best case consists of a conventional supervised finetuning run where \texttt{LLaMA-2} is trained to map $(x_{i}, x_{d}) \rightarrow y$, and thus has full access to the instruction data. This best case performance has no hope of benefiting interpretability,  as there is no way to restrict the interaction between the instruction $x_i$ and data $x_d$ such that we could e.g.\ learn to map the instruction to sparse neurons or directions in latent space. A finetuning run where we ablate the instructions serves as worst case.

We train our system using instruction data from the \texttt{alpaca} dataset \citep{alpaca}, and we report the evaluation perplexity on an unseen split of the data. Table \ref{table:results} outlines the results. Across different layers, the editor consistently performs within the bounds, but there is still a lot of headroom. This suggests that the \texttt{GPT-2} model is able to process the instruction and learn a meaningful (albeit not perfect) manipulation of the frozen \texttt{LLaMA-2} model, across a range of positions in the frozen forward pass where the edit is performed.

\begin{table}[tp]
\centering
\begin{tabular}{@{} l c@{}}
\toprule
                            & eval perplexity (↓ better ) \\ \midrule
Tune w/o instructions & 1.226     \\ \midrule
\texttt{GPT-2} editor (layer 2)  & 1.151     \\
\texttt{GPT-2} editor (layer 10) & 1.148         \\
\texttt{GPT-2} editor (layer 20) & 1.148        \\
\texttt{GPT-2} editor (layer 30) & 1.153        \\ \midrule
Instruction-tune    & 1.026     \\ \bottomrule
\end{tabular}
\caption{\texttt{alpaca} evaluation perplexity for a \texttt{LLaMA-2-7B} processor model, either trained with (ablated) instruction-tuning or using our editor paradigm.}
\label{table:results}
\end{table}

\section{Regularizing Edits for Interpretability}
Our next planned step is to regularize the model edits in a way that promotes interpretability.

For example, if we aim to explain individual neurons, the editor should learn to only manipulate a sparse set of neuron activations. Using our framework, this can be achieved by regularizing the change produced by the edit, given by $\mathcal{E}(x_i, h) - h$, with an L1 loss term such as $\sum_{j=1}^d|\mathcal{E}(x_i, h)_j - h_j|$, where $d$ is the dimension of the hidden representations. Adding this term to our loss intuitively corresponds to learning an editor which can achieve the best manipulation of the frozen model by manipulating a sparse set of activations. Alternatively, edits could be parameterized to directly update model weights, instead of manipulating activations.

The resulting editor performance will give us a macroscopic picture of how conducive these types of edits were. An editor achieving good regularized edit performance would be a valuable resource for downstream interpretability work.

\section*{Limitations}
Model interpretability should be faithful, lest we run the risk of deceiving users and practitioners with plausible, but wrong, model explanations. At inference time, our editor can be exposed to out-of-distribution prompts, causing it to fail. Luckily, because our edits manipulate model behavior, the effectiveness of the edit can be behaviorally verified by the user at inference-time. However, to gain a wide-scale trust in our approach, we will need to verify the faithfulness of edits and explanations resulting from our method on unseen data and concepts, using e.g.\ an explanation verification framework such as CEBaB \citep{abraham2022cebab}.

\section*{Acknowledgements}
KD gratefully acknowledges funding from the FWO Fundamental Research PhD Fellowship (11632223N).

\bibliography{custom}

\begin{thebibliography}{14}
\expandafter\ifx\csname natexlab\endcsname\relax\def\natexlab#1{#1}\fi

\bibitem[{Abraham et~al.(2022)Abraham, D'Oosterlinck, Feder, Gat, Geiger,
  Potts, Reichart, and Wu}]{abraham2022cebab}
Eldar~David Abraham, Karel D'Oosterlinck, Amir Feder, Yair~Ori Gat, Atticus
  Geiger, Christopher Potts, Roi Reichart, and Zhengxuan Wu. 2022.
\newblock \href {https://openreview.net/forum?id=3AbigH4s-ml} {{CEB}ab:
  Estimating the causal effects of real-world concepts on {NLP} model
  behavior}.
\newblock In \emph{Advances in Neural Information Processing Systems}.

\bibitem[{De~Cao et~al.(2022)De~Cao, Schmid, Hupkes, and Titov}]{de2022sparse}
Nicola De~Cao, Leon Schmid, Dieuwke Hupkes, and Ivan Titov. 2022.
\newblock Sparse interventions in language models with differentiable masking.
\newblock In \emph{Proceedings of the Fifth BlackboxNLP Workshop on Analyzing
  and Interpreting Neural Networks for NLP}, pages 16--27.

\bibitem[{Geiger et~al.(2021)Geiger, Lu, Icard, and Potts}]{geiger2021causal}
Atticus Geiger, Hanson Lu, Thomas~F Icard, and Christopher Potts. 2021.
\newblock \href {https://openreview.net/forum?id=RmuXDtjDhG} {Causal
  abstractions of neural networks}.
\newblock In \emph{Advances in Neural Information Processing Systems}.

\bibitem[{Geiger et~al.(2022)Geiger, Wu, Lu, Rozner, Kreiss, Icard, Goodman,
  and Potts}]{pmlr-v162-geiger22a}
Atticus Geiger, Zhengxuan Wu, Hanson Lu, Josh Rozner, Elisa Kreiss, Thomas
  Icard, Noah Goodman, and Christopher Potts. 2022.
\newblock \href {https://proceedings.mlr.press/v162/geiger22a.html} {Inducing
  causal structure for interpretable neural networks}.
\newblock In \emph{Proceedings of the 39th International Conference on Machine
  Learning}, volume 162 of \emph{Proceedings of Machine Learning Research},
  pages 7324--7338. PMLR.

\bibitem[{Leike and Sutskever(2023)}]{superalignment}
Jan Leike and Ilya Sutskever. 2023.
\newblock \href {https://openai.com/blog/introducing-superalignment}
  {Introducing superalignment}.

\bibitem[{Meng et~al.(2022)Meng, Bau, Andonian, and
  Belinkov}]{meng2022locating}
Kevin Meng, David Bau, Alex~J Andonian, and Yonatan Belinkov. 2022.
\newblock \href {https://openreview.net/forum?id=-h6WAS6eE4} {Locating and
  editing factual associations in {GPT}}.
\newblock In \emph{Advances in Neural Information Processing Systems}.

\bibitem[{Mitchell et~al.(2021)Mitchell, Lin, Bosselut, Finn, and
  Manning}]{mitchell2021fast}
Eric Mitchell, Charles Lin, Antoine Bosselut, Chelsea Finn, and Christopher~D
  Manning. 2021.
\newblock Fast model editing at scale.
\newblock In \emph{International Conference on Learning Representations}.

\bibitem[{Mitchell et~al.(2022)Mitchell, Lin, Bosselut, Manning, and
  Finn}]{mitchell2022memory}
Eric Mitchell, Charles Lin, Antoine Bosselut, Christopher~D Manning, and
  Chelsea Finn. 2022.
\newblock Memory-based model editing at scale.
\newblock In \emph{International Conference on Machine Learning}, pages
  15817--15831. PMLR.

\bibitem[{Olsson et~al.(2022)Olsson, Elhage, Nanda, Joseph, DasSarma, Henighan,
  Mann, Askell, Bai, Chen, Conerly, Drain, Ganguli, Hatfield-Dodds, Hernandez,
  Johnston, Jones, Kernion, Lovitt, Ndousse, Amodei, Brown, Clark, Kaplan,
  McCandlish, and Olah}]{olsson2022context}
Catherine Olsson, Nelson Elhage, Neel Nanda, Nicholas Joseph, Nova DasSarma,
  Tom Henighan, Ben Mann, Amanda Askell, Yuntao Bai, Anna Chen, Tom Conerly,
  Dawn Drain, Deep Ganguli, Zac Hatfield-Dodds, Danny Hernandez, Scott
  Johnston, Andy Jones, Jackson Kernion, Liane Lovitt, Kamal Ndousse, Dario
  Amodei, Tom Brown, Jack Clark, Jared Kaplan, Sam McCandlish, and Chris Olah.
  2022.
\newblock In-context learning and induction heads.
\newblock \emph{Transformer Circuits Thread}.
\newblock
  Https://transformer-circuits.pub/2022/in-context-learning-and-induction-heads/index.html.

\bibitem[{Radford et~al.(2019)Radford, Wu, Child, Luan, Amodei, and
  Sutskever}]{radford2019language}
Alec Radford, Jeff Wu, Rewon Child, David Luan, Dario Amodei, and Ilya
  Sutskever. 2019.
\newblock Language models are unsupervised multitask learners.

\bibitem[{Taori et~al.(2023)Taori, Gulrajani, Zhang, Dubois, Li, Guestrin,
  Liang, and Hashimoto}]{alpaca}
Rohan Taori, Ishaan Gulrajani, Tianyi Zhang, Yann Dubois, Xuechen Li, Carlos
  Guestrin, Percy Liang, and Tatsunori~B. Hashimoto. 2023.
\newblock Stanford alpaca: An instruction-following llama model.
\newblock \url{https://github.com/tatsu-lab/stanford_alpaca}.

\bibitem[{Touvron et~al.(2023)Touvron, Martin, Stone, Albert, Almahairi,
  Babaei, Bashlykov, Batra, Bhargava, Bhosale et~al.}]{touvron2023llama}
Hugo Touvron, Louis Martin, Kevin Stone, Peter Albert, Amjad Almahairi, Yasmine
  Babaei, Nikolay Bashlykov, Soumya Batra, Prajjwal Bhargava, Shruti Bhosale,
  et~al. 2023.
\newblock Llama 2: Open foundation and fine-tuned chat models.
\newblock \emph{arXiv preprint arXiv:2307.09288}.

\bibitem[{Vig et~al.(2020)Vig, Gehrmann, Belinkov, Qian, Nevo, Singer, and
  Shieber}]{vig2020causalmediation}
Jesse Vig, Sebastian Gehrmann, Yonatan Belinkov, Sharon Qian, Daniel Nevo,
  Yaron Singer, and Stuart Shieber. 2020.
\newblock \href
  {https://proceedings.neurips.cc/paper_files/paper/2020/file/92650b2e92217715fe312e6fa7b90d82-Paper.pdf}
  {Investigating gender bias in language models using causal mediation
  analysis}.
\newblock In \emph{Advances in Neural Information Processing Systems},
  volume~33, pages 12388--12401. Curran Associates, Inc.

\bibitem[{Wu et~al.(2023)Wu, Geiger, Potts, and Goodman}]{Wu:Geiger:2023:BDAS}
Zhengxuan Wu, Atticus Geiger, Christopher Potts, and Noah~D. Goodman. 2023.
\newblock \href {https://arxiv.org/abs/2305.08809} {Interpretability at scale:
  Identifying causal mechanisms in {Alpaca}}.
\newblock Ms., Stanford University.

\end{thebibliography}
\bibliographystyle{acl_natbib}

\end{document}